\documentclass[11pt,a4paper]{article}
\usepackage[hyperref]{acl2020}
\usepackage{times}
\usepackage{latexsym}
\usepackage{subfigure}

\usepackage{microtype}

\aclfinalcopy 

\usepackage{graphicx}
\usepackage{url}
\usepackage{multirow,makecell}
\usepackage{booktabs}
\usepackage{color}
\usepackage{amsmath,amssymb}

\title{Word-level Textual Adversarial Attacking as Combinatorial Optimization}

\author{Yuan Zang$^{1}$\thanks{\ \ Indicates equal contribution. Yuan developed the method, designed and conducted most experiments; Fanchao formalized the task, designed some experiments and wrote the paper; Chenghao made the original research proposal, performed human evaluation and conducted some experiments.}\hspace{0.3em}, 
Fanchao Qi$^{1*}$, 
Chenghao Yang$^{2*}$\thanks{\ \ Work done during internship at Tsinghua University}\hspace{0.2em}, 
Zhiyuan Liu$^{1}$\thanks{\ \  Corresponding author}\hspace{0.2em}, \\
{\bf Meng Zhang$^{3}$,
Qun Liu$^{3}$,
Maosong Sun$^{1}$}\\
$^{1}$Department of Computer Science and Technology, Tsinghua University \\
Institute for Artificial Intelligence, Tsinghua University \\
Beijing National Research Center for
Information Science and Technology\\
$^{2}$Columbia University \quad
$^{3}$Huawei Noah's Ark Lab\\
{\tt \{zangy17,qfc17\}@mails.tsinghua.edu.cn, yangalan1996@gmail.com}\\
{\tt \{liuzy,sms\}@tsinghua.edu.cn, \{zhangmeng92,qun.liu\}@huawei.com}
}

\date{}

\begin{document}
\maketitle
\begin{abstract}
Adversarial attacks are carried out to reveal the vulnerability of deep neural networks. 
Textual adversarial attacking is challenging because text is discrete and a small perturbation can bring significant change to the original input.
Word-level attacking, which can be regarded as a combinatorial optimization problem, is a well-studied class of textual attack methods.
However, existing word-level attack models are far from perfect, largely because unsuitable search space reduction methods and inefficient optimization algorithms are employed.
In this paper, 
we propose a novel attack model, which incorporates the sememe-based word substitution method and particle swarm optimization-based search algorithm to solve the two problems separately.
We conduct exhaustive experiments to evaluate our attack model by attacking BiLSTM and BERT on three benchmark datasets. 
Experimental results demonstrate that our model consistently achieves much higher attack success rates and crafts more high-quality adversarial examples as compared to baseline methods.
Also, further experiments show our model has higher transferability and can bring more robustness enhancement to victim models by adversarial training.
All the code and data of this paper can be obtained on \url{https://github.com/thunlp/SememePSO-Attack}.
\end{abstract}

\section{Introduction}
\interfootnotelinepenalty=10000

Adversarial attacks use \textit{adversarial examples} \citep{szegedy2014intriguing,goodfellow2015explaining}, which are maliciously crafted by perturbing the original input, to fool the deep neural networks (DNNs). 
Extensive studies have demonstrated that DNNs are vulnerable to adversarial attacks, e.g., minor modification to highly poisonous phrases can easily deceive Google's toxic comment detection systems \citep{hosseini2017deceiving}.
From another perspective, adversarial attacks are also used to improve robustness and interpretability of DNNs \citep{wallace2019universal}.
In the field of natural language processing (NLP) which widely employs DNNs, 
practical systems such as spam filtering \citep{stringhini2010detecting} and malware detection \citep{kolter2006learning} have been broadly used, but at the same time the concerns about their security are growing.
Therefore, the research on textual adversarial attacks becomes increasingly important.


\begin{figure}[!t]
    \centering
    \includegraphics[width=\linewidth]{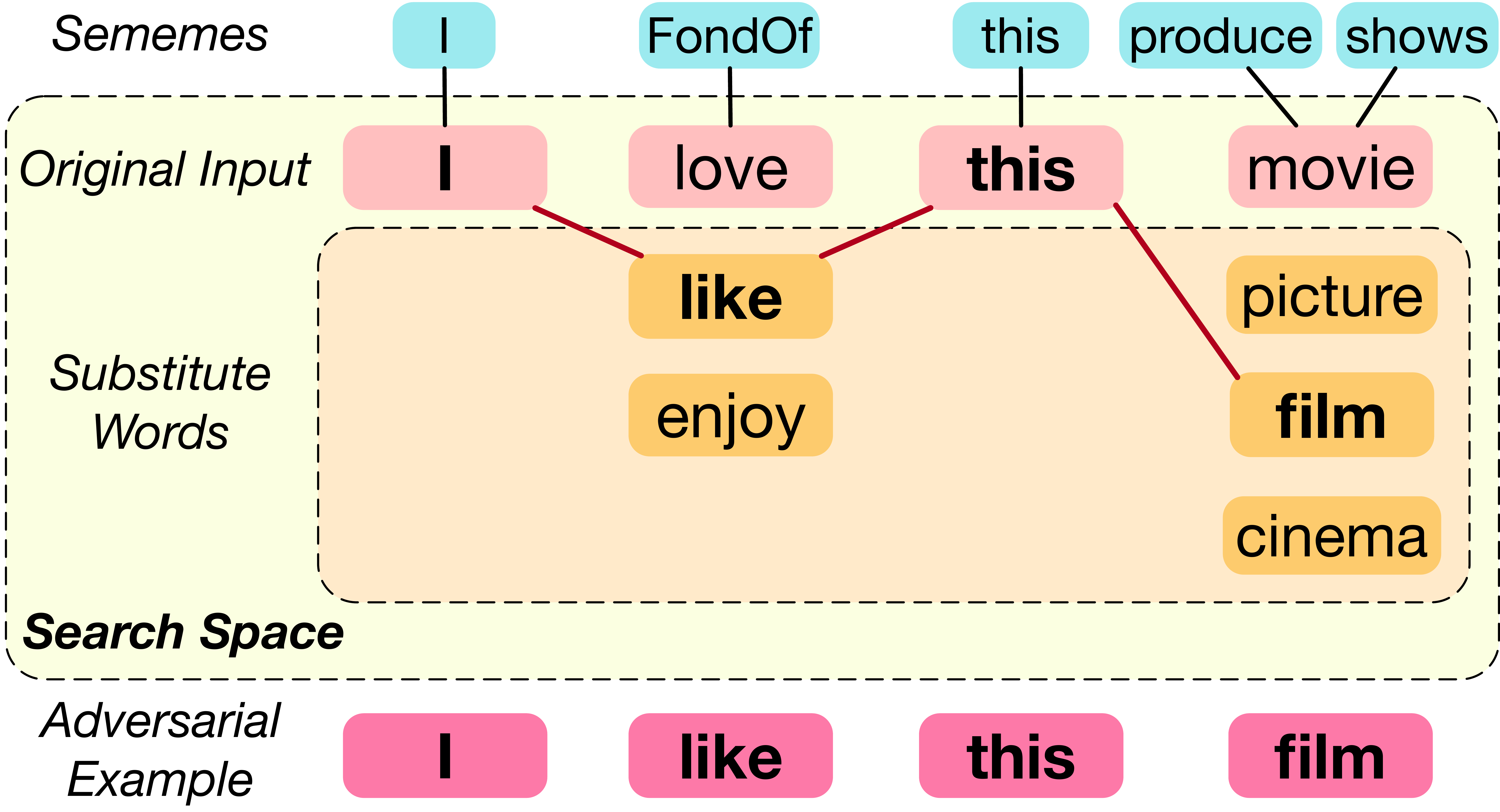}
    \caption{An example showing search space reduction with sememe-based word substitution and adversarial example search in word-level adversarial attacks.}
    \label{fig:example}
\end{figure}

Textual adversarial attacking is challenging.
Different from images, a truly imperceptible perturbation on text is almost impossible because of its discrete nature. 
Even a slightest character-level perturbation can either (1) change the meaning and, worse still, the true label of the original input, or (2) break its grammaticality and naturality. 
Unfortunately, the change of true label will make the adversarial attack \textit{invalid}.
For example, supposing an adversary changes ``she'' to ``he'' in an input sentence to attack a gender identification model, although the victim model alters its prediction result, this is not a valid attack.
And the adversarial examples with broken grammaticality and naturality (i.e., poor quality) can be easily defended \citep{pruthi2019combating}.

Various textual adversarial attack models have been proposed \citep{wang2019survey}, ranging from character-level flipping \citep{ebrahimi2018hotflip} to sentence-level paraphrasing \citep{iyyer2018adversarial}.
Among them, word-level attack models, mostly word substitution-based models, perform comparatively well on both attack efficiency and adversarial example quality \citep{wang2019natural}. 

Word-level adversarial attacking is actually a problem of \textit{combinatorial optimization} \citep{wolsey1999integer},
as its goal is to craft adversarial examples 
which can successfully fool the victim model using a limited vocabulary. 
In this paper, as shown in Figure \ref{fig:example}, we 
break this combinatorial optimization problem down into two steps including (1) reducing search space and (2) searching for adversarial examples.

The first step is aimed at excluding invalid or low-quality potential adversarial examples and retaining the valid ones with good grammaticality and naturality.
The most common manner is to pick some candidate substitutes for each word in the original input and use their combinations as the reduced discrete search space.
However, existing attack models either disregard this step \citep{papernot2016crafting} or adopt unsatisfactory substitution methods that do not perform well in the trade-off between quality and quantity of the retained adversarial examples \citep{alzantot2018generating,ren2019generating}. 
The second step is supposed to find adversarial examples that can successfully fool the victim model in the reduced search space. 
Previous studies have explored diverse search algorithms including gradient descent \citep{papernot2016crafting}, genetic algorithm \citep{alzantot2018generating} and greedy algorithm \citep{ren2019generating}.
Some of them like gradient descent only work in the white-box setting where full knowledge of the victim model is required.
In real situations, however, we usually have no access to the internal structures of victim models.
As for the other black-box algorithms, they are not efficient and effective enough in searching for adversarial examples.



These problems negatively affect the overall attack performance of existing word-level adversarial attacking.
To solve the problems, we propose a novel black-box word-level adversarial attack model, which reforms both the two steps.
In the first step, we design a word substitution method based on \textit{sememes}, the minimum semantic units, which can retain more potential valid adversarial examples with high quality.
In the second step, we present a search algorithm based on \textit{particle swarm optimization} \citep{eberhart1995particle}, which is very efficient and performs better in finding adversarial examples.
We conduct exhaustive experiments 
to evaluate our model. 
Experimental results show that, compared with baseline models, our model not only achieves the highest attack success rate (e.g., 100\% when attacking BiLSTM on IMDB) but also possesses the best adversarial example quality and comparable attack validity.
We also conduct decomposition analyses to manifest the advantages of the two parts of our model separately.
Finally, we demonstrate that our model has the highest transferability and can bring the most robustness improvement to victim models by adversarial training.

\section{Background}
In this section, we first briefly introduce sememes, and then we give an overview of the classical particle swarm optimization algorithm.

\subsection{Sememes}

In linguistics, a sememe is defined as the minimum semantic unit of human languages \citep{bloomfield1926set}. 
The meaning of a word can be represented by the composition of its sememes. 

In the field of NLP, sememe knowledge bases are built to utilize sememes in practical applications, where sememes are generally regarded as semantic labels of words (as shown in Figure \ref{fig:example}).
HowNet \citep{zhendong2006hownet} is the most well-known one.
It annotates over one hundred thousand English and Chinese words with a predefined sets of about 2,000 sememes. 
Its sememe annotations are sense-level, i.e., each sense of a (polysemous) word is annotated with sememes separately. 
With the help of 
HowNet,
sememes have been successfully applied to many NLP tasks including word representation learning \citep{niu2017improved}, sentiment analysis \citep{fu2013multi}, 
semantic composition \citep{qi2019modeling}, sequence modeling \citep{qin2019enhancing}, reverse dictionary \citep{zhang2019multi}, etc.

\subsection{Particle Swarm Optimization}
Inspired by the social behaviors like bird flocking, particle swarm optimization (PSO) is a kind of metaheuristic population-based evolutionary computation paradigms \citep{eberhart1995particle}.
It has been proved effective in solving the optimization problems such as 
image classification \citep{omran2004image}, part-of-speech tagging \citep{silva2012biopos} and text clustering \citep{cagnina2014efficient}.
Empirical studies have proven it is more efficient than some other optimization algorithms like the genetic algorithm \citep{hassan2005comparison}.

PSO exploits a population of interacting individuals to iteratively search for the optimal solution in the specific space. 
The population is called a \textit{swarm} and the individuals are called \textit{particles}. 
Each particle has a \textit{position} in the search space and moves with an adaptable \textit{velocity}. 

Formally, when searching in a $D$-dimensional continuous space $\mathcal{S} \subseteq \mathbb{R}^D$ with a swarm containing $N$ particles, the position and velocity of each particle can be represented by $\mathbf{x}^n \in \mathcal{S}$ and $\mathbf{v}^n\in \mathbb{R}^D$ respectively, $n\in \{1,\cdots,N\}$.
Next we describe the PSO algorithm step by step.

(1) \textbf{Initialize.}
At the very beginning, each particle is randomly initialized with a position $\mathbf{x}^n$ in the search space and a velocity $\mathbf{v}^n$.
Each dimension of the initial velocity $v^n_{d} \in [-V_{max},V_{max}]$, $d \in \{1,\cdots,D\}$.

(2) \textbf{Record.}
Each position in the search space corresponds to an optimization score.
The position a particle has reached with the highest optimization score is recorded as its \textit{individual best position}.
The best position among the individual best positions of all the particles
is recorded as the \textit{global best position}.

(3) \textbf{Terminate.}
If current global best position has achieved the desired optimization score, the algorithm terminates and outputs the global best position as the search result.

(4) \textbf{Update.}
Otherwise, the velocity and position of each particle are updated according to its current position and individual best position together with the global best position. 
The updating formulae 
are
\begin{equation}
\begin{aligned}
v^n_d &= \omega v^n_d +c_1\times{r_1}\times (p^n_d - x^n_d) \\
& \qquad\quad + c_2\times{r_2}\times(p^g_d-x^n_d),\\
x^n_d & = x^n_d+v^n_d,\\
\end{aligned}
\end{equation}
where $\omega$ is 
the inertia weight, 
$p^n_{d}$ and $p^g_{d}$ are the $d$-th dimensions of the $n$-th particle's individual best position and the global best position respectively,
$c_1$ and $c_2$ are acceleration coefficients which are positive constants and control how fast the particle moves towards its individual best position and the global best position,
and $r_1$ and $r_2$ are random coefficients.
After updating, the algorithm goes back to the \textbf{Record} step.


\section{Methodology}

In this section, we detail our word-level adversarial attack model.
It incorporates two parts, namely the sememe-based word substitution method and PSO-based adversarial example search algorithm.

\subsection{Sememe-based Word Substitution Method}

The sememes of a word are supposed to accurately depict the meaning of the word \cite{zhendong2006hownet}. 
Therefore, the words with the same sememe annotations should have the same meanings, and they can serve as the substitutes for each other.

Compared with other word substitution methods, mostly including word embedding-based \citep{sato2018interpretable}, language model-based \citep{zhang2019generating} and synonym-based methods \citep{samanta2017towards,ren2019generating}, the sememe-based word substitution method can achieve a better trade-off between quality and quantity of substitute words. 

For one thing, although the word embedding and language model-based substitution methods can find as many substitute words as we want simply by relaxing the restrictions on embedding distance and language model prediction score, they inevitably introduce many inappropriate and low-quality substitutes, such as antonyms and semantically related but not similar words, into adversarial examples which might break the semantics, grammaticality and naturality of original input.
In contrast, the sememe-based and, of course, the synonym-based substitution methods does not have this problem.

For another, compared with the synonym-based method, the sememe-based method can find more substitute words and, in turn, retain more potential adversarial examples, 
because HowNet annotates sememes for all kinds of words.
The synonym-based method, however, depends on thesauri like WordNet \citep{miller1995wordnet}, which provide no synonyms for many words like proper nouns and the number of a word's synonyms is very limited.
An empirical comparison of different word substitution methods is given in Section \ref{sec:decom}.
In our sememe-based word substitution method, to preserve grammaticality, 
we only substitute content words\footnote{Content words are the words that carry meanings and consist mostly of nouns, verbs, adjectives and adverbs.} and restrict the substitutes to having the same part-of-speech tags as the original words. 
Considering polysemy, a word $w$ can be substituted by another word $w_*$ only if one of $w$'s senses has the same sememe annotations as one of $w_*$'s senses.
When making substitutions, we conduct lemmatization to enable more substitutions and delemmatization to avoid introducing grammatical mistakes.

\subsection{PSO-based Adversarial Example Search Algorithm}
\label{sec:PSOsearch}

Before presenting our algorithm, we first 
explain what the concepts in the original PSO algorithm correspond to in the adversarial example search problem.

Different from original PSO, 
the search space of word-level adversarial example search is discrete. 
A \textit{position} in the search space corresponds to a sentence (or an adversarial example), and each dimension of a position corresponds to a word. 
Formally, 
$\mathbf{x}^n=w^n_1 \cdots w^n_d \cdots w^n_D, w^n_d \in \mathbb{V}(w^o_d)$, 
where $D$ is the length (word number) of the original input, $w^o_d$ is the $d$-th word in the original input, and $\mathbb{V}(w^o_d)$ is composed of $w^o_d$ and its substitutes.

The optimization score of a position is the \textit{target label}'s prediction probability given by the victim model, where the target label is the desired classification result for an adversarial attack.
Taking a binary classification task as an example, if the true label of the original input is ``positive'', the target label is ``negative'', and vice versa.
In addition, a particle's \textit{velocity} now relates to the position change probability, i.e., $v^n_{d}$ determines how probable $w^n_{d}$ is substituted by another word.

Next we describe our algorithm step by step. 

First, for the \textbf{Initialize} step, since we expect the adversarial examples to differ from the original input as little as possible, we do not make random initialization.  
Instead, we randomly substitute one word of the original input to determine the initial position of a particle. 
This operation is actually the \textit{mutation} of genetic algorithm, which has also been employed in some studies on discrete PSO \citep{higashi2003particle}.
We repeat mutation $N$ times to initialize the positions of $N$ particles.
Each dimension of each particle's velocity is randomly initialized between $-V_{max}$ and $V_{max}$.

For the \textbf{Record} step, our algorithm keeps the same as the original PSO algorithm. 
For the \textbf{Terminate} step, the termination condition is the victim model predicts the target label for any of current adversarial examples. 


\begin{table*}[!t]
\resizebox{\linewidth}{!}{
\begin{tabular}{ccccccccc}
\toprule
Dataset & Task & \#Class & Avg. \#W & Train  & Dev   & Test  & BiLSTM \%ACC & BERT \%ACC \\ 
\midrule
IMDB    & Sentiment Analysis & 2     & 234    & 25000  & 0  & 25000 & 89.10      & 90.76    \\
SST-2     & Sentiment Analysis & 2     & 17     & 6920   & 872   & 1821  & 83.75      & 90.28    \\
SNLI    & NLI & 3     & 8      & 550152 & 10000 & 10000 & 84.43      & 89.58   \\ 
\bottomrule
\end{tabular}
}
\caption{Details of datasets and their accuracy results of victim models. 
``\#Class'' means the number of classifications.
``Avg. \#W'' signifies the average sentence length (number of words). 
``Train'', ``Val'' and ``Test'' denote the instance numbers of the training, validation and test sets respectively.
``BiLSTM \%ACC'' and ``BERT \%ACC'' means the classification accuracy of BiLSTM and BERT.}
\label{tab:dataset}
\end{table*}

For the \textbf{Update} step, considering the discreteness of search space, 
we follow \citet{kennedy1997discrete} to adapt the updating formula of velocity to
\begin{equation}
\begin{aligned}
v^n_d=\omega v^n_d+(1-\omega)&\times[\mathcal{I}(p^n_d,x^n_d)+\mathcal{I}(p^g_d,x^n_d)],
\end{aligned}
\end{equation}
where $\omega$ is still the inertia weight, and $\mathcal{I}(a,b)$ is defined as
\begin{equation}
\mathcal{I}(a,b)=\left\{
\begin{aligned}
1 & , & a=b, \\
-1 & , & a\neq b.
\end{aligned}
\right.
\end{equation}

Following \citet{shi1998parameter}, we let the inertia weight decrease with the increase of numbers of iteration times, aiming to make the particles highly dynamic to explore more positions in the early stage and gather around the best positions quickly in the final stage.
Specifically, 
\begin{equation}
\omega=(\omega_{max}-\omega_{min})\times\frac{T-t}{T}+\omega_{min},
\end{equation}
where $0<\omega_{min}<\omega_{max}<1$, and $T$ and $t$ are the maximum and current numbers of iteration times.

The updating of positions also needs to be adjusted to the discrete search space. 
Inspired by \citet{kennedy1997discrete}, instead of making addition, we adopt a probabilistic method to update the position of a particle to the best positions.
We design two-step position updating. 
In the first step, a new movement probability $P_i$ is introduced, with which a particle determines whether it moves to its \textit{individual best position} as a whole.
Once a particle decides to move, the change of each dimension of its position depends on the same dimension of its velocity, specifically with the probability of $\mathrm{sigmoid}(v^n_d)$.
No matter whether a particle has moved towards its individual best position or not, it would be processed in the second step.
In the second step, each particle determines whether to move to the \textit{global best position} with another movement probability $P_g$. 
And the change of each position dimension also relies on $\mathrm{sigmoid}(v^n_d)$.
$P_i$ and $P_g$ vary with iteration to enhance search efficiency by adjusting the balance between local and global search, i.e., encouraging particles to explore more space around their individual best positions in the early stage and search for better position around the global best position in the final stage. 
Formally,
\begin{equation}
\begin{aligned}
    P_i&=P_{max}-\frac{t}{T}\times(P_{max}-P_{min}),\\
    P_g&=P_{min}+\frac{t}{T}\times(P_{max}-P_{min}),
\end{aligned}
\end{equation}
where $0<P_{min}<P_{max}<1$.

Besides, to enhance the search in unexplored space, we apply \textbf{mutation} to each particle after the update step. To avoid excessive modification, mutation is conducted with the probability
\begin{equation}
P_m(\mathbf{x}^n)=\max\left(0, 1-k \frac {\mathcal{E}(\mathbf{x}^n,\mathbf{x}^o)}{D}\right),
\label{eq:Pm}
\end{equation}
where $k$ is a positive constant, $\mathbf{x}^o$ represents the original input, and $\mathcal{E}$ measures the word-level edit distance (number of different words between two sentences).
$\frac {\mathcal{E}(\mathbf{x}^n,\mathbf{x}^o)}{D}$ is defined as the \textit{modification rate} of an adversarial example.
After mutation, the algorithm returns to the \textbf{Record} step.

\section{Experiments}
In this section, we conduct comprehensive experiments to evaluate our attack model on the tasks of sentiment analysis and natural language inference.



\subsection{Datasets and Victim Models}
For sentiment analysis, we choose two benchmark datasets including IMDB \citep{maas2011learning} and SST-2 \citep{socher2013recursive}.
Both of them are binary sentiment classification datasets.
But the average sentence length of SST-2 (17 words) is much shorter than that of IMDB (234 words), which renders attacks on SST-2 more challenging.
For natural language inference (NLI), we use the popular Stanford Natural Language Inference (SNLI) dataset \citep{bowman2015large}. 
Each instance in SNLI comprises a premise-hypothesis sentence pair and is labelled one of three relations including entailment, contradiction and neutral.

As for victim models, we choose two widely used universal sentence encoding models, namely bidirectional LSTM (BiLSTM) with max pooling \citep{conneau2017supervised} and BERT\textsubscript{BASE} (BERT) \citep{devlin2019bert}.
For BiLSTM, its hidden states are 128-dimensional, and it uses 300-dimensional pre-trained GloVe \citep{pennington2014glove} word embeddings.
Details of the datasets and the classification accuracy results of the victim models are listed in Table \ref{tab:dataset}.



\subsection{Baseline Methods} 
We select two recent open-source word-level adversarial attack models as the baselines, which are typical and involve different search space reduction methods (step 1) and search algorithms (step 2).

The first baseline method \citep{alzantot2018generating} uses the combination of restrictions on word embedding distance and language model prediction score to reduce search space. 
As for search algorithm, it adopts genetic algorithm, another popular metaheuristic population-based evolutionary algorithm.
We use ``\textbf{Embedding/LM+Genetic}'' to denote this baseline method.

The second baseline \citep{ren2019generating} chooses synonyms from WordNet \citep{miller1995wordnet} as substitutes and designs a saliency-based greedy algorithm as the search algorithm.
We call this method ``\textbf{Synonym+Greedy}''. 
This baseline model is very similar to another attack model TextFooler \citep{jin2019bert}, which has extra semantic similarity checking when searching adversarial examples. But we find the former performs better in almost all experiments, and thus we only select the former as a baseline for comparison.

In addition, to conduct decomposition analyses of different methods in the two steps separately, we combine different search space reduction methods (Embedding/LM, Synonym and our sememe-based substitution method (Sememe)), and search algorithms (Genetic, Greedy and our PSO).

\subsection{Experimental Settings}
For our PSO, $V_{max}$ is set to 1, $\omega_{max}$ and $\omega_{min}$ are set to 0.8 and 0.2, $P_{max}$ and $P_{min}$ are also set to 0.8 and 0.2, and $k$ in Equation \eqref{eq:Pm} is set to 2.
All these hyper-parameters have been tuned on the validation set.
For the baselines, we use their recommended hyper-parameter settings.
For the two population-based search algorithms Genetic and PSO, we set the maximum number of iteration times ($T$ in Section \ref{sec:PSOsearch}) to 20 and the population size ($N$ in Section \ref{sec:PSOsearch}) to 60, which are the same as \citet{alzantot2018generating}.




\begin{table}[!t]
\resizebox{1.02\linewidth}{!}{
\begin{tabular}{ccc}
\toprule
Metrics     & Evaluation Method & Better? \\
\midrule
Success Rate       & Auto         & Higher       \\
Validity           & Human (Valid Attack Rate)        & Higher       \\
Modification Rate  & Auto         & Lower        \\
Grammaticality  & Auto (Error Increase Rate)  & Lower        \\
Fluency         & Auto (Perplexity)        & Lower        \\
Naturality         & Human (Naturality Score)          & Higher \\
\bottomrule
\end{tabular}
}
\caption{Details of evaluation metrics. 
``Auto'' and ``Human'' represent automatic and human evaluations respectively.
``Higher'' and ``Lower'' mean the higher/lower the metric, the better a model performs.}
\label{tab:metric}
\end{table}

\subsection{Evaluation Metrics}
To improve evaluation efficiency, we randomly sample $1,000$ correctly classified instances from the test sets of the three datasets as the original input to be perturbed. 
For SNLI, only the hypotheses are perturbed.
Following \citet{alzantot2018generating}, 
we restrict the length of the original input to $10$-$100$,
exclude the out-of-vocabulary words from the substitute sets, and discard the adversarial examples with modification rates higher than 25\%.


\begin{table*}[!t]
\centering
\resizebox{.7\linewidth}{!}{
\begin{tabular}{cc||rrr|rrr}
\toprule
 {\multirow{2}{*}{\makecell[c]{Word Substitution \\ Method}}} & {\multirow{2}{*}{\makecell[c]{Search \\ Algorithm}}} & \multicolumn{3}{c|}{BiLSTM}                        & \multicolumn{3}{c}{BERT}                         \\\cline{3-8}
                           &                                            & \multicolumn{1}{c}{IMDB}            & \multicolumn{1}{c}{SST-2}            & \multicolumn{1}{c|}{SNLI}           & \multicolumn{1}{c}{IMDB}           & \multicolumn{1}{c}{SST-2}            & \multicolumn{1}{c}{SNLI}           \\ \hline

\multirow{3}{*}{Embedding/LM}                            & Genetic                                                       & 86.90           & 67.70          & 44.40          & 87.50          & 66.20          & 44.30          \\
    & Greedy                                                       & 80.90           & 69.00          & 47.70          & 62.50          & 56.20          & 42.40 \\
                                              
                                               & PSO                                                          & {96.90}           & {78.50}          & {50.90}          & {93.60}          & {74.40}         & {53.10}                   \\\hline
\multirow{3}{*}{Synonym}                          & Genetic                                                         & 95.50           & 73.00          & 51.40          & 92.90          & 78.40          & 56.00 \\
                                               & Greedy                                                         & 87.20           & 73.30          & 57.70          & 73.00          & 64.60          & 52.70          \\
                                              &PSO & {98.70}           & {79.20}          & {61.80}          & {96.20}          & {80.90}         & {62.60}
                                                        \\\hline
\multirow{3}{*}{Sememe}                           & Genetic                                                       & 96.90           & 78.50          & 50.90          & 93.60          & 74.40          & 53.10          \\
                                               & Greedy                                                          & 95.20           & 87.70          & 70.40          & 80.50          & 74.80          & 66.30           \\
                                               & PSO                                                          & \textbf{100.00} & \textbf{93.80} & \textbf{73.40} & \textbf{98.70} & \textbf{91.20} & \textbf{78.90} \\ \bottomrule
\end{tabular}
}
\caption{The attack success rates (\%) of different attack models.}
\label{tab:main}
\end{table*}

\begin{table*}[!t]
\centering

\resizebox{.95\linewidth}{!}{
\begin{tabular}{cc||rrr|rrr|rrr}
\toprule
\multirow{2}{*}{\makecell[c]{Victim\\Model}} & \multirow{2}{*}{Attack Model} & \multicolumn{3}{c|}{IMDB}                       & \multicolumn{3}{c|}{SST-2}                          & \multicolumn{3}{c}{SNLI}                          \\ \cline{3-11}
                              &                               & \multicolumn{1}{c}{\%M}  & \multicolumn{1}{c}{\%I}       & \multicolumn{1}{c|}{PPL}            & \multicolumn{1}{c}{\%M}   & \multicolumn{1}{c}{\%I}        & \multicolumn{1}{c|}{PPL}             & \multicolumn{1}{c}{\%M}    & \multicolumn{1}{c}{\%I}        & \multicolumn{1}{c}{PPL}             \\ \hline
\multirow{3}{*}{BiLSTM}       & Embedding/LM+Genetic                  & 9.76          & 5.49          & 124.20         & 12.03         & 7.08          & 319.98          & 13.31          & 14.12          & 235.20          \\                   
                              & Synonym+Greedy                  & 6.47          & 4.49          & 115.31         & 10.25        & 4.65          & 317.27          & 12.32         & 21.37          & 311.04          \\
                              & Sememe+PSO                    & \textbf{3.71}  & \textbf{1.44} & \textbf{88.98} & \textbf{9.06}  & \textbf{3.17} & \textbf{276.53} & \textbf{11.72}  & \textbf{11.08} & \textbf{222.40} \\ \hline
\multirow{3}{*}{BERT}         & Embedding/LM+Genetic                  & 7.41        & 4.22          & 106.12         & 10.41         & 5.09          & 314.22          & 13.04          & 15.09          & 225.92          \\                   
                              & Synonym+Greedy                  & 4.49          & 4.48          & 98.60          & 8.51          & 4.11          & 316.30          & \textbf{11.60}         & 11.65          & 285.00          \\
                              & Sememe+PSO                    & \textbf{3.69}  & \textbf{1.57} & \textbf{90.74} & \textbf{8.24}  & \textbf{2.03} & \textbf{289.94} & 11.72  & \textbf{10.14} & \textbf{223.22} \\ \bottomrule

\end{tabular}
}
\caption{Automatic evaluation results of adversarial example quality. 
``\%M'', ``\%I'' and ``PPL'' indicate the modification rate, grammatical error increase rate and language model perplexity respectively.}
\label{tab:auto}
\vspace{-0.5em}
\end{table*}

\begin{table}[!t]
\centering
\resizebox{1.02\linewidth}{!}{
\begin{tabular}{cc|cc}
\toprule
Victim            & Attack Model & \%Valid & NatScore \\ \hline
N/A& Original Input & 90.0 & 2.30 \\ \hline
\multirow{3}{*}{BiLSTM} & Embedding/LM+Genetic           & 65.5      & 2.205       \\
                        & Synonym+Greedy         & \textbf{72.0}      & 2.190       \\
                        & Sememe+PSO          & 70.5      & \textbf{2.210}       \\ \hline
\multirow{3}{*}{BERT}   & Embedding/LM+Genetic           & \textbf{74.5}      & 2.165       \\
                        & Synonym+Greedy         & 66.5      & 2.165       \\
                        & Sememe+PSO          & 72.0      & \textbf{2.180}       \\ \bottomrule
\end{tabular}
}
\caption{Human evaluation results of attack validity and adversarial example naturality on SST-2, where the second row additionally lists the evaluation results of original input.
``\%Valid'' refers to the percentage of valid attacks.
``NatScore'' is the average naturality score of adversarial examples.
}
\label{tab:human}
\end{table}

We evaluate the performance of attack models including their attack success rates, attack validity and the quality of adversarial examples. 
The details of our evaluation metrics are listed in Table \ref{tab:metric}.

(1) The attack success rate is defined as the percentage of the attacks which craft an adversarial example to make the victim model predict the target label. 
(2) The attack validity is measured by the percentage of valid attacks to successful attacks, where the adversarial examples crafted by valid attacks have the same true labels as the original input.
(3) For the quality of adversarial examples, we divide it into four parts including modification rate, grammaticality, fluency and naturality.
\textit{Grammaticality} is measured by the increase rate of grammatical error numbers of adversarial examples compared with the original input, where we use LanguageTool
\footnote{\url{https://www.languagetool.org}} to obtain the grammatical error number of a sentence.
We utilize the language model perplexity (PPL) to measure the \textit{fluency} with the help of GPT-2 \citep{radford2019language}.
The \textit{naturality} reflects whether an adversarial example is natural and indistinguishable from human-written text.

We evaluate attack validity and adversarial example naturality only on SST-2 by human evaluation with the help of Amazon Mechanical Turk\footnote{\url{https://www.mturk.com}}. 
We randomly sample $200$ adversarial examples,
and ask the annotators to make a binary sentiment classification and give a naturality score (1, 2 or 3, higher better) for each adversarial example and original input.
More annotation details are given in Appendix \ref{sec:human_eval}.




\subsection{Attack Performance}

\paragraph{Attack Success Rate}
The attack success rate results of all the models are listed in Table \ref{tab:main}.
We observe that our attack model (Sememe+PSO) achieves the highest attack success rates on all the three datasets (especially the harder SST-2 and SNLI) and two victim models, proving the superiority of our model over baselines.
It attacks BiLSTM/BERT on IMDB with a notably 100.00\%/98.70\% success rate, which clearly demonstrates the vulnerability of DNNs.
By comparing three word substitution methods (search space reduction methods) and three search algorithms, we find Sememe and PSO consistently outperform their counterparts.
Further decomposition analyses are given in a later section.

\paragraph{Validity and Adversarial Example Quality}
We evaluate the attack validity and adversarial example quality of our model together with the two baseline methods (Embedding/LM+Genetic and Synonym+Greedy).
The results of automatic and human evaluations are displayed in Table \ref{tab:auto} and \ref{tab:human} respectively.\footnote{Automatic evaluation results of adversarial example quality of all the combination models are shown in Appendix \ref{sec:auto_eval}.}
Note that the human evaluations including attack validity and adversarial example naturality are conducted on SST-2 only.
We find that in terms of automatic evaluations of adversarial example quality, including modification rate, grammaticality and fluency, our model consistently outperforms the two baselines on whichever victim model and dataset.
As for attack validity and adversarial example naturality, our Sememe+PSO model obtains a slightly higher overall performance than the two baselines. 
But its adversarial examples are still inferior to original human-authored input, especially in terms of validity (label consistency).

We conduct Student's \textit{t}-tests to further measure the difference between the human evaluation results of different models, where the statistical significance threshold of \textit{p}-value is set to $0.05$. 
We find that neither of the differences of attack validity and adversarial example naturality between different models are significant.
In addition, the adversarial examples of any attack model have significantly worse label consistency (validity) than the original input, but possesses similar naturality.
More details of statistical significance test are given in Appendix \ref{sec:human_stat_significance}.


For Embedding/LM, relaxing the restrictions on embedding distance and language model prediction score can improve its attack success rate but sacrifices attack validity. 
To make a specific comparison, we adjust the hyper-parameters of Embedding/LM+Genetic\footnote{The detailed hyper-parameter settings are given in Appendix \ref{sec:hyper_adjust}.} to increase its attack success rates to $96.90\%$, $90.30\%$, $58.00\%$,  $93.50\%$, $83.50\%$ and $62.90\%$ respectively on attacking the two victim models on the three datasets (in the same order as Table \ref{tab:main}). 
Nonetheless, its attack validity rates against BiLSTM and BERT on SST-2 dramatically fall to $59.5\%$ and $56.5\%$.
In contrast, ours are $70.5\%$ and $72.0\%$, and their differences are significant according to the results of significance tests in Appendix \ref{sec:human_stat_significance}.

\subsection{Decomposition Analyses}
\label{sec:decom}
In this section, we conduct detailed decomposition analyses of different word substitution methods (search space reduction methods) and different search algorithms, aiming to further demonstrate the advantages of our sememe-based word substitution method and PSO-based search algorithm.

\paragraph{Word Substitution Method}
Table \ref{tab:number} lists the average number of substitutes provided by different word substitution methods on the three datasets. 
It shows Sememe can find much more substitutes than the other two counterparts, which explains the high attack success rates of the models incorporating Sememe.
Besides, we give a real case from SST-2 in Table \ref{tab:example} which lists substitutes found by the three methods.
We observe that Embedding/LM find many improper substitutes, Synonym cannot find any substitute because the original word ``pie'' has no synonyms in WordNet, and only Sememe finds many appropriate substitutes.



\begin{table}[!t]
\centering
\resizebox{.95\linewidth}{!}{
\begin{tabular}{c|rrr}
\toprule
Word Substitution Method & \multicolumn{1}{c}{IMDB}  & \multicolumn{1}{c}{SST-2} &    \multicolumn{1}{c}{SNLI}  \\ \hline
Embedding/LM             & 3.44  & 3.27  & 3.42  \\
Synonym               & 3.55  & 3.08  & 3.14  \\
Sememe                & \textbf{13.92} & \textbf{10.97} & \textbf{12.87} \\\bottomrule
\end{tabular}
}
\caption{The average number of substitutes provided by different word substitution methods.}
\label{tab:number}
\end{table}

\begin{table}[!t]
\centering
\setlength{\belowcaptionskip}{-0.2cm}
\resizebox{1.02\linewidth}{!}{
\begin{tabular}{c|c|c}
\toprule
\multicolumn{3}{l}{
\begin{tabular}[c]{@{}c@{}}
She breaks the {\color{green}pie} dish and screams out that she is not handicapped.
\end{tabular}}
\\ 
\hline
Embedding/LM& Synonym& Sememe \\ 
\hline
\begin{tabular}[c]{@{}c@{}}{\color{red}tart}, {\color{red}pizza}, {\color{red}apple},\\ shoemaker, {\color{red}cake} \\ {\color{red}cheesecake}\end{tabular} &
\begin{tabular}[c]{@{}c@{}}None\end{tabular} &
\begin{tabular}[c]{@{}c@{}}{\color{red}cheese}, {\color{red}popcorn}, {\color{red}ham}, {\color{red}cream},\\ {\color{red}break}, {\color{red}cake}, {\color{red}pizza}, {\color{red}chocolate}, \\ 
and 55 more 
\end{tabular}
\\
\bottomrule
\end{tabular}
}
\caption{A real case showing the substitutes found by three word substitution methods, where the original word is colored green and appropriate substitutes are colored red.}
\label{tab:example}
\end{table}

\begin{figure}[!t]
\centering
\setlength{\belowcaptionskip}{-0.4cm}
\includegraphics[width=\linewidth]{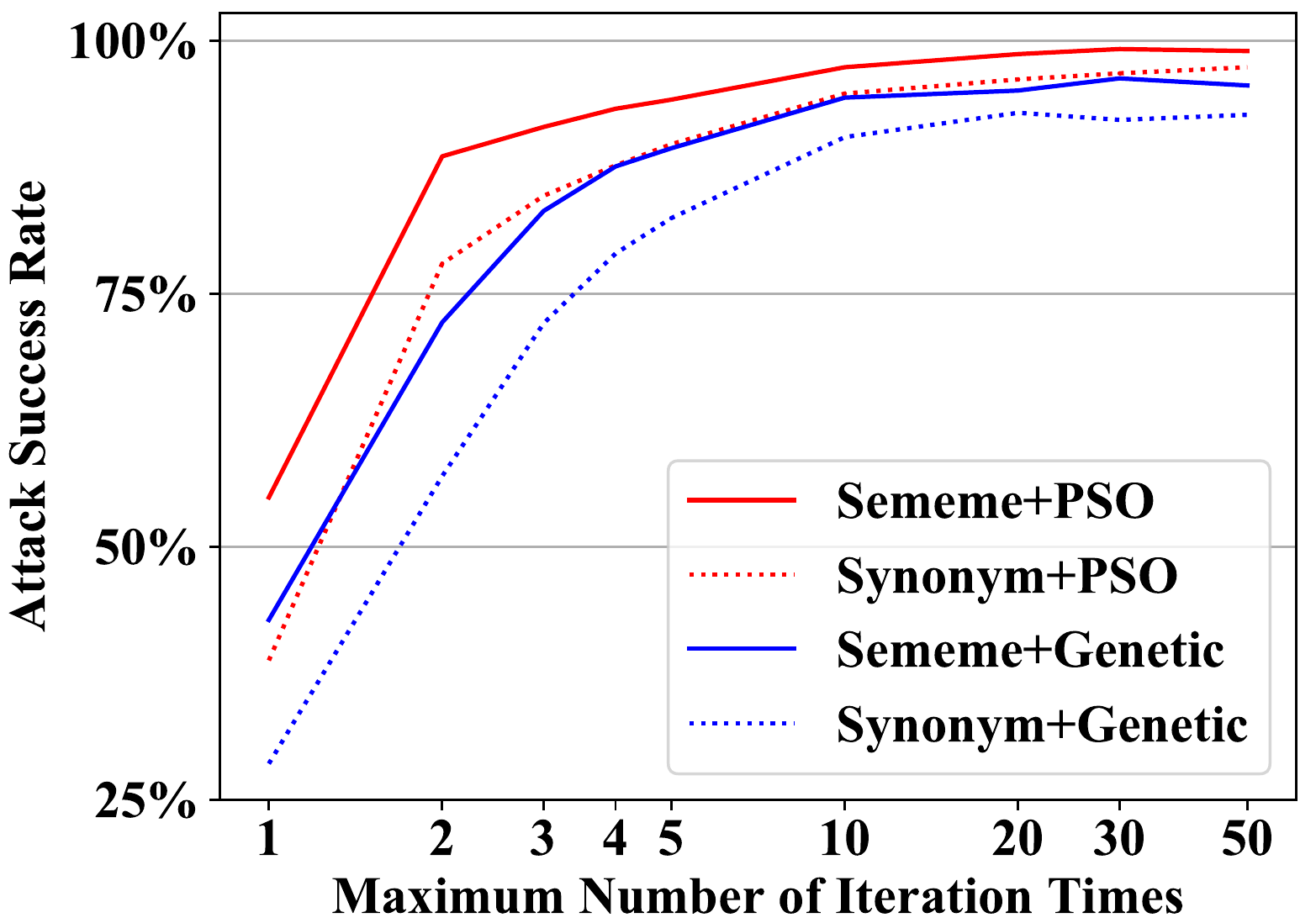}
\caption{Attack success rates of different models with different maximum numbers of iteration times. The x-coordinate is in log-2 scale.}
\label{fig:com1}
\end{figure}

\begin{figure}[!t]
\centering
\includegraphics[width=\linewidth]{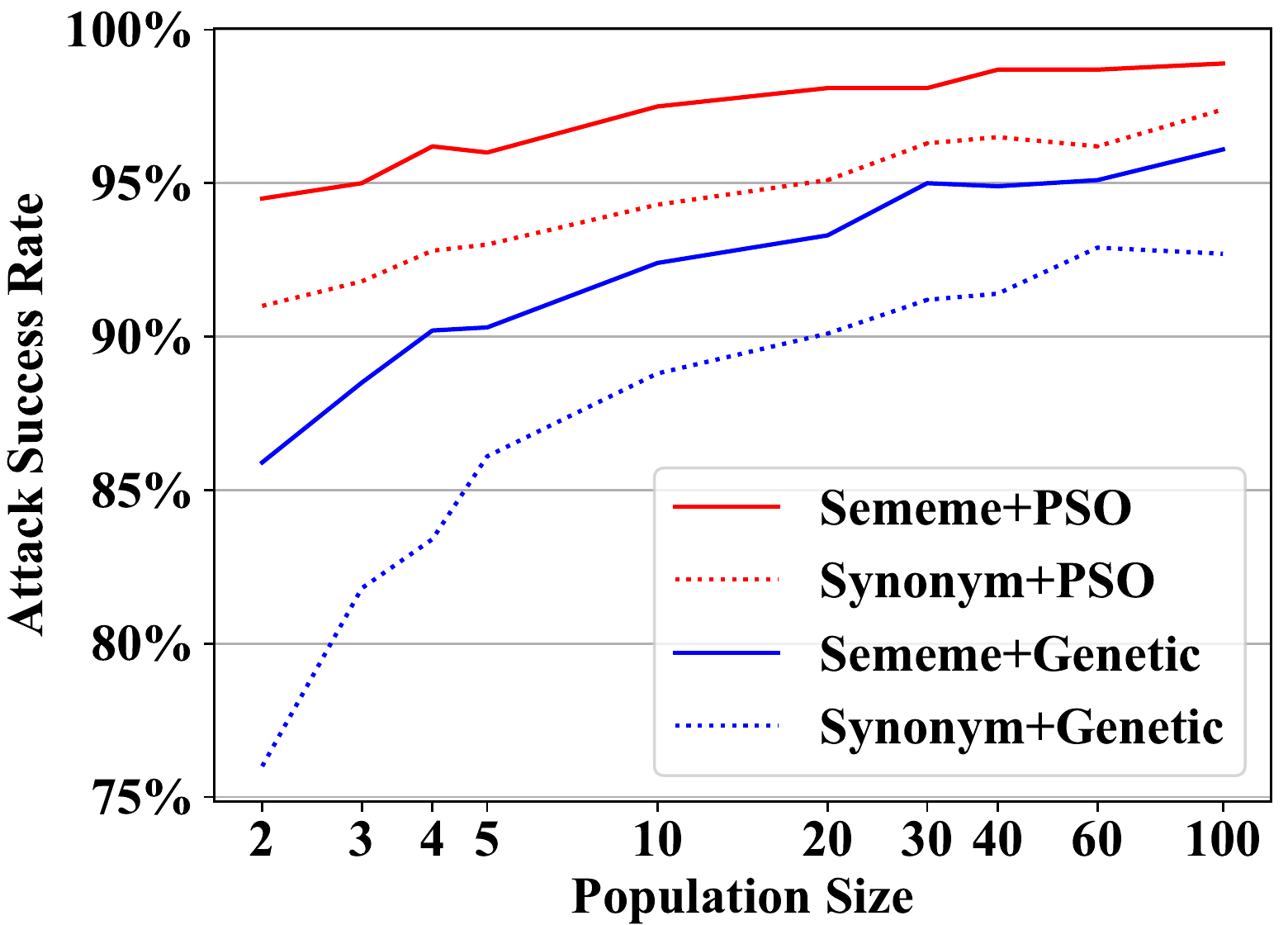}
\caption{Attack success rates of different models with population sizes. The x-coordinate is in log-2 scale.}
\label{fig:com2}
\end{figure}

\paragraph{Search Algorithm}
We compare the two population-based search algorithms Genetic and PSO by changing two important hyper-parameters, namely the maximum number of iteration times $T$ and the population size $N$.
The results of attack success rate are shown in Figure \ref{fig:com1} and \ref{fig:com2}.
From the two figures, we find our PSO outperforms Genetic consistently, especially in the setting with severe restrictions on maximum number of iteration times and population size, which highlights the efficiency of PSO.


\begin{table}[!t]
\centering
\setlength{\belowcaptionskip}{-0.3cm}
\resizebox{1.02\linewidth}{!}{
\begin{tabular}{cc|ccc}
\toprule
Transfer            & Attack Model & IMDB  & SST-2 & SNLI  \\ \hline
 BiLSTM & Embedding/LM+Genetic           & 81.93 & 70.61 & 61.26 \\
   $\Downarrow$                     & Synonym+Greedy         & 77.29 & 64.94 & 65.34 \\
   BERT                     & Sememe+PSO          & \textbf{75.80} & \textbf{64.71} & \textbf{59.54} \\ \hline
  BERT & Embedding/LM+Genetic           & 86.63 & 65.71 & 49.66 \\
    $\Downarrow$                    & Synonym+Greedy         & 81.64 & 58.67 & \textbf{45.16} \\
      BiLSTM                  & Sememe+PSO           & \textbf{78.42} & \textbf{58.11} & 46.89 \\ \bottomrule
\end{tabular}
}
\caption{The classification accuracy of transferred adversarial examples on the three datasets. 
Lower accuracy reflects higher transferability.}
\label{tab:trans}
\end{table}

\subsection{Transferability}
The transferability of adversarial examples reflects whether an attack model can attack a DNN model without any access to it \citep{kurakin2016adversarial}.
It has been widely used as an important evaluation metric in adversarial attacks.
We evaluate the transferability of adversarial examples by using BiLSTM to classify the adversarial examples crafted for attacking BERT, and vice versa.
Table \ref{tab:trans} shows the classification accuracy results of transferred adversarial examples. 
Note that lower accuracy signifies higher transferability.
The lower the accuracy is, the higher the transferability is.
We find compared with the two baselines, our Sememe+PSO crafts adversarial examples with overall higher transferability.

\subsection{Adversarial Training}
Adversarial training is proposed to improve the robustness of victim models by adding adversarial examples to the training set \citep{goodfellow2015explaining}. 
In this experiment, for each attack model, we craft 692 adversarial examples (10\% of the original training set size) by using it to attack BiLSTM on the \textit{training} set of SST-2. 
Then we add the adversarial examples to the training set and re-train a BiLSTM. 
We re-evaluate its robustness by calculating the attack success rates of different attack models.
Table \ref{tab:adv_train} lists the results of adversarial training. 
Note larger attack success rate decrease signifies greater robustness improvement.
We find that adversarial training can improve the robustness of victim models indeed, and our Sememe+PSO model brings greater robustness improvement than the two baselines, even when the attack models are exactly themselves.\footnote{For instance, using Embedding/LM+Genetic in adversarial training to defend its attack declines the attack success rate by $2.60\%$ while using our Sememe+PSO model declines by $3.53\%$.}
From the perspective of attacking, our Sememe+PSO model is still more threatening than others even under the defense of adversarial training.

We also manually select 692 \textit{valid} adversarial examples generated by Sememe+PSO to conduct adversarial training, which leads to even greater robustness improvement (last column of Table \ref{tab:adv_train}).
The results show that adversarial example validity has big influence on adversarial training effect.


\section{Related Work}
Existing textual adversarial attack models can be classified into three categories according to the perturbation levels of their adversarial examples.

Sentence-level attacks include adding distracting sentences \citep{jia2017adversarial}, paraphrasing \citep{iyyer2018adversarial,ribeiro2018semantically} and performing perturbations in the continuous latent semantic space \citep{zhao2018generating}.
Adversarial examples crafted by these methods usually have profoundly different forms from original input and their validity are not guaranteed.

Character-level attacks are mainly random character manipulations including swap, substitution, deletion, insertion and repeating \citep{belinkov2018synthetic,gao2018black,hosseini2017deceiving}.
In addition, gradient-based character substitution methods have also been explored, with the help of one-hot character embeddings \citep{ebrahimi2018hotflip} or visual character embeddings \cite{eger2019text}.
Although character-level attacks can achieve high success rates, they break the grammaticality and naturality of original input and can be easily defended 
\citep{pruthi2019combating}.

\begin{table}[!t]

\centering
\setlength{\belowcaptionskip}{-0.5cm}
\resizebox{1.02\linewidth}{!}{
\begin{tabular}{c|rrrrr}
\toprule
Att \textbackslash Adv.T & \multicolumn{1}{c}{None} & \multicolumn{1}{c}{E/L+G} & \multicolumn{1}{c}{Syn+G} & \multicolumn{1}{c}{Sem+P} &\multicolumn{1}{c}{Sem+P*} \\ 
\hline
E/L+G          & 67.70 & -2.60 & -0.60  & -3.53 & -5.10 \\
Syn+G         & 73.30 & -2.67 & -3.50 & -3.13 & -3.53 \\
Sem+P          & 93.80 & -1.07 & 0.03  & -2.93 & -4.33 \\ \bottomrule
\end{tabular}
}
\caption{
The attack success rates of different attack models when attacking BiLSTM on SST-2 and their decrements brought by adversarial training.
``Att'' and ``Adv.T'' denote  ``Attack Model'' and ``Adversarial Training''.
E/L+G, Syn+G and Sem+P represent Embedding/LM+Genetic, Synonym+Greedy and our Sememe+PSO, respectively. 
``Sem+P*'' denotes only using the \textit{valid} adversarial examples generated by Sememe+PSO in adversarial training.
}
\label{tab:adv_train}
\end{table}

As for word-level attacks, following our two-step modeling, their adversarial example space reduction methods (step 1) involve 
using word embeddings \citep{sato2018interpretable} or language model \citep{zhang2019generating} to filter words,
selecting synonyms 
as substitutes \citep{samanta2017towards,ren2019generating,jin2019bert}, 
and their combinations \citep{alzantot2018generating,glockner2018breaking}. 
The search algorithms (step 2) include gradient descent \citep{papernot2016crafting,sato2018interpretable,gong2018adversarial}, genetic algorithm \citep{alzantot2018generating}, Metropolis-Hastings sampling \citep{zhang2019generating}, saliency-based greedy algorithm \citep{liang2018deep,ren2019generating,jin2019bert}.
In comparison, our model adopts new methods in both steps which are more powerful.

\section{Conclusion and Future Work}
In this paper, we propose a novel word-level attack model comprising the sememe-based word substitution method and particle swarm optimization-based search algorithm.
We conduct extensive experiments to demonstrate the superiority of our model in terms of attack success rate, adversarial example quality, transferability and robustness improvement to victim models by adversarial training.
In the future, we will try to increase the robustness gains of adversarial training and consider utilizing sememes in adversarial defense model. 

\section*{Acknowledgments}
This work is supported by the National Key Research and Development Program of China (No. 2018YFB1004503) and the National Natural Science Foundation of China (NSFC No. 61732008, 61772302). 
We also thank the anonymous reviewers for their valuable comments and suggestions.

\bibliography{main}
\bibliographystyle{acl_natbib}

\appendix

\begin{table*}[!t]
\centering
\resizebox{.95\linewidth}{!}{
\begin{tabular}{c||c|c||rrr|rrr|rrr}
\toprule
\multirow{2}{*}{\makecell[c]{Victim\\Model}} & {\multirow{2}{*}{\makecell[c]{Word Substitution \\ Method}}} & {\multirow{2}{*}{\makecell[c]{Search \\ Algorithm}}}  & \multicolumn{3}{c|}{IMDB}                       & \multicolumn{3}{c|}{SST-2}                          & \multicolumn{3}{c}{SNLI}                          \\ \cline{4-12}
                              &             &                  & \multicolumn{1}{c}{\%M}  & \multicolumn{1}{c}{\%I}       & \multicolumn{1}{c|}{PPL}            & \multicolumn{1}{c}{\%M}   & \multicolumn{1}{c}{\%I}        & \multicolumn{1}{c|}{PPL}             & \multicolumn{1}{c}{\%M}    & \multicolumn{1}{c}{\%I}        & \multicolumn{1}{c}{PPL}             \\ \hline
\multirow{9}{*}{BiLSTM}       & \multirow{3}{*}{Embedding/LM} & Genetic                  & 9.76          & 5.49          & 124.20         & 12.03         & 7.08          & 319.98          & 13.31          & 14.12          & 235.20          \\ \cline{3-12}
 & & Greedy                  & 7.84          & 5.23          & 112.84        & 10.63         & 3.71          & 287.45          & 12.60          & \textbf{9.42}          & \textbf{205.50}          \\ \cline{3-12}
 & & PSO                  & 7.00          & 8.07          & 113.99         & 12.78         & 6.68          & 339.46          & 14.82          & 10.82          & 255.69          \\ \cline{2-12}
& \multirow{3}{*}{Synonym} & Genetic                  & 7.60          & 6.07         & 137.51         & 11.35        & 5.32          & 357.19          & 12.60         & 24.78          & 283.95          \\ \cline{3-12}
                              &  & Greedy                  & 6.47          & 4.49          & 115.31         & 10.25        & 4.65          & 317.27          & 12.32         & 21.37          & 311.04          \\ \cline{3-12}
                              &  & PSO                  & 5.42          & 3.45          & 109.27         & 10.55        & 5.12          & 331.96          & 12.56         & 20.83          & 307.51          \\ \cline{2-12}
                               & \multirow{3}{*}{Sememe} & Genetic                 & 5.30          & 2.55          & 105.24         & 10.04        & 3.48          & 298.49          & \textbf{11.30}         & 11.64          & 205.61          \\ \cline{3-12}
                              &  & Greedy                  & 4.89          &1.80          & 97.49         & 9.36        & \textbf{2.79}          & \textbf{276.53}         & 12.11         & 10.95          & 218.72         \\ \cline{3-12}
                              &  & PSO                  & \textbf{3.71}          & \textbf{1.44}          & \textbf{88.98}         & \textbf{9.06}        & 3.17          & \textbf{276.53}          & 11.72         & 11.08          & 222.40          \\ \hline \hline
\multirow{9}{*}{BERT}         & \multirow{3}{*}{Embedding/LM} & Genetic                  & 7.41        & 4.22          & 106.12         & 10.41         & 5.09          & 314.22          & 13.04          & 15.09          & 225.92          \\ \cline{3-12} 
 &  & Greedy                  & 5.53        & 4.45          & 97.21        & 9.23         & 3.04          & \textbf{276.42}         & 11.80          & 13.73         & \textbf{206.46}         \\  \cline{3-12}
  &  & PSO                  & 5.97        & 7.98          & 101.66        & 11.64         & 6.70          & 343.89          & 14.22          & 14.43          & 245.95          \\  \cline{2-12}
   & \multirow{3}{*}{Synonym} & Genetic                  & 5.72        & 5.59         & 114.57         & 9.62         & 4.62          & 353.05          & 13.09          & 13.01          & 311.14          \\  \cline{3-12}
    &  & Greedy                  & 4.49        & 4.48          & 98.60         & 8.51         & 4.12          & 316.30          & 11.60          & 11.65          & 285.00          \\  \cline{3-12}
     &  & PSO                  & 4.63        & 4.33          & 100.81         & 9.20         & 4.72          & 337.82          & 12.99         & 13.32          & 302.83          \\  \cline{2-12}
      & \multirow{3}{*}{Sememe} & Genetic                  & 4.27        & 1.62          & 97.86         & 8.34         & 2.05          & 292.16          & 11.59          & 8.84          & 217.75          \\  \cline{3-12}
       &  & Greedy                  & 3.97        & 1.79          & 92.31         & \textbf{8.14}         & 2.21          & 279.35          & \textbf{10.09}          & \textbf{7.81}          & 207.71          \\  \cline{3-12}
        &  & PSO                 & \textbf{3.69}        &\textbf{1.57}          & \textbf{90.74}        & 8.24         & \textbf{2.03}          & 289.94         & 11.73          & 10.14         & 223.22          \\  \bottomrule

\end{tabular}
}
\caption{Automatic evaluation results of adversarial example quality. 
``\%M'', ``\%I'' and ``PPL'' indicate the modification rate, grammatical error increase rate and language model perplexity respectively.}
\label{tab:auto_full}
\vspace{-0.5em}
\end{table*}

\section{Human Evaluation Details}
\label{sec:human_eval}
For each adversarial example and original input, we ask three workers to choose a sentiment label from ``Positive'' and ``Negative'' for it and annotate its naturality score from $\{1,2,3\}$, which indicates ``Machine generated'', ``Not sure'' and ``Human written'' respectively. We get the final sentiment labels of adversarial examples by voting. 
For example, if two workers annotate an example as ``Positive'' and one worker annotates it as ``Negative'', we record its annotated label as ``Positive''. 
We obtain the validity rate by calculating the percentage of the adversarial examples which are annotated with the same sentiment labels as corresponding original sentences.
For each adversarial example, we use the average of the naturality scores given by three workers as its final naturality score.

\section{Automatic Evaluation Results of Adversarial Example Quality}
\label{sec:auto_eval}
We present the automatic evaluation results of adversarial example quality of all the combination models in Table \ref{tab:auto_full}.
We can find that Sememe and PSO obtain higher overall adversarial example quality than other word substitution methods and adversarial example search algorithms, respectively.

\section{Adjustment of Hyper-parameters of Embedding/LM+Genetic}
\label{sec:hyper_adjust}
The word substitution strategy Embedding/LM has three hyper-parameters: the number of the nearest words $\mathrm{N}$, the euclidean distance threshold of word embeddings $\mathrm{\delta}$ and the number of words retained by the language model filtering $\mathrm{K}$.
For original Embedding/LM+Genetic, $\mathrm{N}$ = 8, $\mathrm{\delta = 0.5}$ and $\mathrm{K = 4}$, which are the same as \citet{alzantot2018generating}. 
To increase the attack success rates, we change these hyper-parameters to $\mathrm{N = 20}$, $\mathrm{\delta} = 1$ and $\mathrm{K = 10}$. 



\section{Statistical Significance of Human Evaluation Results}
\label{sec:human_stat_significance}
We conduct Student's \textit{t}-tests to measure the statistical significance between the difference of human evaluation results of different models.
The results of attack validity and adversarial example naturality are shown in Table \ref{tab:valid_t_test} and \ref{tab:nat_t_test}, respectively.
``Embedding/LM+Genetic*'' refers to the Embedding/LM+Genetic model with adjusted hyper-parameters.

\begin{table*}[t]
\centering
\resizebox{.95\linewidth}{!}{
\begin{tabular}{c||cc||lcc}
\toprule
Victim Model             & Model 1 & Model 2 & \textit{p}-value & Significance    & Conclusion     \\ \hline
\multirow{7}{*}{Bi-LSTM} & Sememe+PSO                  & Embedding/LM+Genetic        & 0.14                     & Not Significant & =              \\ \cline{2-6}
                         & Sememe+PSO                  & Synonym+Greedy              & 0.37                     & Not Significant & =              \\ \cline{2-6}
                         & Sememe+PSO                  &
                         Embedding/LM+Genetic*     &   0.01 & Significant & \textgreater{} \\ \cline{2-6}
                         & Original Input              & Embedding/LM+Genetic        & 9.52e-10                 & Significant     & \textgreater{} \\ \cline{2-6}
                         & Original Input              & Synonym+Greedy              & 1.78e-6                  & Significant     & \textgreater{} \\ \cline{2-6}
                         & Original Input              & Sememe+PSO                  & 3.55e-7                  & Significant     & \textgreater{} \\ \cline{2-6}
                         & Original Input              & Embedding/LM+Genetic*                  & 2.42e-13                  & Significant     & \textgreater{} \\ \hline \hline
\multirow{7}{*}{BERT}    & Sememe+PSO                  & Embedding/LM+Genetic        & 0.29                     & Not Significant & =              \\ \cline{2-6}
                         & Sememe+PSO                  & Synonym+Greedy              & 0.12                     & Not Significant & =              \\ \cline{2-6}
                         & Sememe+PSO                  &
                         Embedding/LM+Genetic*     &   5.86e-4 & Significant & \textgreater{} \\ \cline{2-6}
                         & Original Input              & Embedding/LM+Genetic        & 2.19e-5                  & Significant     & \textgreater{} \\ \cline{2-6}
                         & Original Input              & Synonym+Greedy              & 3.33e-9                  & Significant     & \textgreater{} \\ \cline{2-6}
                         & Original Input              & Sememe+PSO                  & 1.78e-6                  & Significant     & \textgreater{} \\ \cline{2-6}
                         & Original Input              & Embedding/LM+Genetic*                  & 2.30e-15                  & Significant     & \textgreater{} \\  \bottomrule
\end{tabular}
}
\caption{The Student's \textit{t}-test results of attack validity of different models, where ``='' means ``Model 1'' performs as well as ``Model 2'' and ``\textgreater{}'' means ``Model 1'' performs better than ``Model 2''.}
\label{tab:valid_t_test}
\end{table*}

\begin{table*}[t]
\centering
\resizebox{.95\linewidth}{!}{
\begin{tabular}{c||cc||lcc}
\toprule
Victim Model             & Model 1 & Model 2 & \textit{p}-value & Significance    & Conclusion     \\ \hline
\multirow{5}{*}{Bi-LSTM} & Sememe+PSO                  & Embedding/LM+Genetic        & 0.48                     & Not Significant & =              \\ \cline{2-6}
                         & Sememe+PSO                  & Synonym+Greedy              & 0.41                     & Not Significant & =              \\ \cline{2-6}
                         & Human Authored              & Embedding/LM+Genetic        & 0.14                 & Not Significant     & = \\ \cline{2-6}
                         & Human Authored              & Synonym+Greedy              & 0.10                 & Not Significant     & = \\ \cline{2-6}
                         & Human Authored              & Sememe+PSO                  & 0.15                  & Not Significant     & = \\ \hline \hline
\multirow{5}{*}{BERT}    & Sememe+PSO                  & Embedding/LM+Genetic        & 0.31                     & Not Significant & =              \\ \cline{2-6}
                         & Sememe+PSO                  & Synonym+Greedy              & 0.31                     & Not Significant & =              \\ \cline{2-6}
                         & Human Authored              & Embedding/LM+Genetic        & 0.06                 & Not Significant     & = \\ \cline{2-6}
                         & Human Authored              & Synonym+Greedy              & 0.06                  & Not Significant     & = \\ \cline{2-6}
                         & Human Authored              & Sememe+PSO                  & 0.08                  & Not Significant     & = \\  \bottomrule
\end{tabular}
}
\caption{The Student's \textit{t}-test results of adversarial example naturality of different models, where ``='' means ``Model 1'' performs as well as ``Model 2''.}
\label{tab:nat_t_test}
\end{table*}

\section{Case Study}
\label{sec:case}
We display some adversarial examples generated by the baseline attack models and our attack model on IMDB, SST-2 and SNLI in Table \ref{tab:imdb}, \ref{tab:sst} and \ref{tab:snli} respectively.

\begin{table*}[!t]\resizebox{1.02\linewidth}{!}{
    \centering
    \begin{tabular}{l}
         \toprule[2pt]
         \multicolumn{1}{c}{IMDB Example 1}\\ 
         \hline
         \textbf{Original Input} (Prediction = \textbf{Positive}) \\ \hline
         In my {\color{green}opinion} this is the {\color{green}best} oliver stone flick {\color{green}probably} more because of  influence than anything else.\\ {\color{green}Full} of {\color{green}dread} from the first moment to its dark ending. \\ 
         \midrule[1.5pt]
        \textbf{Embedding/LM+Genetic}  (Prediction = \textbf{Negative}) \\
         \hline
         In my {\color{red}view} this is the {\color{red}higher} oliver stone flick {\color{red}presumably} more because of influence than anything else. \\{\color{red}Total} of {\color{red}anxiety} from the first moment to its dark ending. \\
         \midrule[1.5pt]
         \textbf{Synonym+Greedy}  (Prediction = \textbf{Negative}) \\ \hline
         In my opinion this {\color{red}embody} the {\color{red}respectable} oliver stone flick probably more because of influence than \\anything else. {\color{red}Broad} of dread from the first moment to its dark ending. \\ 
         \midrule[1.5pt]
         \textbf{Sememe+PSO}  (Prediction = \textbf{Negative})\\ \hline
         In my opinion this is the bestest oliver stone flick probably more because of influence than anything else. \\{\color{red}Ample} of dread from the first moment to its dark ending.\\ 
    
         \toprule[2pt]
         \multicolumn{1}{c}{IMDB Example 2}\\ \hline
         \textbf{Original Input} (Prediction = \textbf{Negative}) \\ \hline
         One of the {\color{green}worst} films of it's genre. The only bright spots {\color{green}were} lee {\color{green}showing} some of the {\color{green}sparkle}\\ she would {\color{green}later} bring to the time tunnel and batman. \\ \midrule[1.5pt]
         \textbf{Embedding/LM+Genetic} (Prediction = \textbf{Positive}) \\\hline
         One of the {\color{red}biggest} films of it's genre. The only {\color{red}glittering} spots were lee showing some of the sparkle\\ she would {\color{red}afterwards} bring to the time tunnel and batman.  \\\midrule[1.5pt]
         \textbf{Synonym+Greedy} (Prediction = \textbf{Positive}) \\ \hline
         One of the {\color{red}tough} films of it's genre. The only bright spots {\color{red}follow} lee {\color{red}present} some of the {\color{red}spark}\\ she would later bring to the time tunnel and batman. \\ \midrule[1.5pt]
         \textbf{Sememe+PSO} (Prediction = \textbf{Positive})\\ \hline
         One of the {\color{red}seediest} films of it's genre. The only {\color{red}shimmering} spots were lee showing some of the sparkle\\ she would later bring to the time tunnel and batman.\\ 
         \bottomrule[2pt]
    \end{tabular}
    }
    \caption{Adversarial examples generated by two baseline methods and our model on IMDB.}
    \label{tab:imdb}
\end{table*}
\begin{table*}[!t]
    \centering
    \resizebox{1.02\linewidth}{!}{
    \begin{tabular}{l}
         \toprule[2pt]
         \multicolumn{1}{c}{SST-2 Example 1}\\ \hline
         \textbf{Original Input} (Prediction = \textbf{Positive}) \\ \hline
         Some actors have so much charisma that you 'd be {\color{green}happy} to listen to them {\color{green}reading} the phone book. \\ \midrule[1.5pt]
         \textbf{Embedding/LM+Genetic} (Prediction = \textbf{Negative}) \\\hline
         Some actors have so much charisma that you 'd be {\color{red}cheery} to listen to them reading the phone book. \\\midrule[1.5pt]
         \textbf{Synonym+Greedy} (Prediction = \textbf{Negative}) \\ \hline
         Some actors have so much charisma that you 'd be happy to listen to them {\color{red}take} the phone book. \\ \midrule[1.5pt]
         \textbf{Sememe+PSO} (Prediction = \textbf{Negative})\\ \hline
         Some actors have so much charisma that you 'd be {\color{red}jovial} to listen to them reading the phone book.\\ \toprule[2pt]

         \multicolumn{1}{c}{SST-2 Example 2}\\ \hline
         Original Sentence (Prediction = \textbf{Negative}) \\ \hline
         The movie 's biggest is its complete and utter {\color{green}lack} of {\color{green}tension}. \\ \midrule[1.5pt]
         \textbf{Embedding/LM+Genetic} (Prediction = \textbf{Positive}) \\\hline
         The movie 's biggest is its complete and utter {\color{red}absence} of {\color{red}stress}.  \\\midrule[1.5pt]
         \textbf{Synonym+Greedy} (Prediction = \textbf{Positive}) \\ \hline
         The movie 's great is its complete and utter {\color{red}want} of tension. \\ \midrule[1.5pt]
         \textbf{Sememe+PSO} (Prediction = \textbf{Positive})\\ \hline
         The movie 's biggest is its complete and utter {\color{red}dearth} of tension.\\ 
         \bottomrule[2pt]
    \end{tabular}
    }
    \caption{Adversarial examples generated by two baseline methods and our model on SST-2.}
    \label{tab:sst}
\end{table*}
\begin{table*}[!t]
    \centering
    \resizebox{1.02\linewidth}{!}{
    \begin{tabular}{l}
         \toprule[2pt]
         \multicolumn{1}{c}{SNLI Example 1}\\ \hline
         \textbf{Premise}: A smiling bride sits in a swing with her smiling groom standing behind her posing for the male \\photographer while a boy holding a bottled drink and another boy wearing a green shirt observe . \\ \midrule[1.5pt]
         \textbf{Original Input}(Prediction = \textbf{Entailment}) \\ \hline
         Two {\color{green}boys} {\color{green}look} on as a {\color{green}married} couple {\color{green}get} their pictures taken. \\ \midrule[1.5pt]
         \textbf{Embedding/LM+Genetic}  (Prediction = \textbf{Contradiction}) \\\hline
          Two {\color{red}man} {\color{red}stare} on as a {\color{red}wedding} couple get their pictures taken.\\\midrule[1.5pt]
         \textbf{Synonym+Greedy}  (Prediction = \textbf{Contradiction}) \\ \hline
         Two boys look on as a married couple {\color{red}puzzle} their pictures taken.\\ \midrule[1.5pt]
         \textbf{Sememe+PSO} (Prediction = \textbf{Contradiction})\\ \hline
         Two boys {\color{red}stare} on as a {\color{red}wedding} couple get their pictures taken.\\

         \toprule[2pt]
         \multicolumn{1}{c}{SNLI Example 2}\\ \hline
         \textbf{Premise}: A dog with a purple leash is held by a woman wearing white shoes .\\ \midrule[1.5pt]
         \textbf{Original Input} (Prediction = \textbf{Entailment}) \\ \hline
         A {\color{green}man} is holding a leash on someone {\color{green}else} {\color{green}dog}. \\ \midrule[1.5pt]
         \textbf{Embedding/LM+Genetic} (Prediction = \textbf{Contradiction}) \\\hline
         A man is holding a leash on someone {\color{red}further} dog.\\\midrule[1.5pt]
         \textbf{Synonym+Greedy} (Prediction = \textbf{Contradiction}) \\ \hline
         A {\color{red}humans} is holding a leash on someone else dog.\\ \midrule[1.5pt]
         \textbf{Sememe+PSO} (Prediction = \textbf{Contradiction})\\ \hline
         A man is holding a leash on someone else {\color{red}canine}.\\ 
         \bottomrule[2pt]
    \end{tabular}
    }
    \caption{Adversarial examples generated by two baseline methods and our model on SNLI.}
    \label{tab:snli}
\end{table*}

\end{document}